\documentclass[conference]{IEEEtran}
\usepackage{hyperref}

\usepackage{cite}
\usepackage{amsmath,amssymb,amsfonts}
\usepackage{graphicx}
\usepackage{textcomp}
\usepackage{xcolor}
\usepackage{algorithm,algpseudocode}

\def\BibTeX{{\rm B\kern-.05em{\sc i\kern-.025em b}\kern-.08em
    T\kern-.1667em\lower.7ex\hbox{E}\kern-.125emX}}
\begin{document}

\title{Mapping Image Transformations Onto Pixel Processor Arrays\\
}

\author{Laurie Bose \kern 5mm Piotr Dudek$^{1}$\\
$^{1}$University of Manchester, Manchester, United Kingdom\\}

\maketitle

\begin{abstract}
Pixel Processor Arrays (PPA) present a new vision sensor/processor architecture consisting of a SIMD array of processor elements, each capable of light capture, storage, processing and local communication. Such a device allows visual data to be efficiently stored and manipulated directly upon the focal plane, but also demands the invention of new approaches and algorithms, suitable for the massively-parallel fine-grain processor arrays. In this paper we demonstrate how various image transformations, including shearing, rotation and scaling, can be performed directly upon a PPA. The implementation details are presented using the SCAMP-5 vision chip, that contains a 256x256 pixel-parallel array. Our approaches for performing the image transformations efficiently exploit the parallel computation in a cellular processor array, minimizing the number of SIMD instructions required. These fundamental image transformations are vital building blocks for many visual tasks. This paper aims to serve as a reference for future PPA research while demonstrating the flexibility of PPA architectures.  
\end{abstract}


\section{Introduction}
\begin{figure}
\centering
\includegraphics[width=1\linewidth]{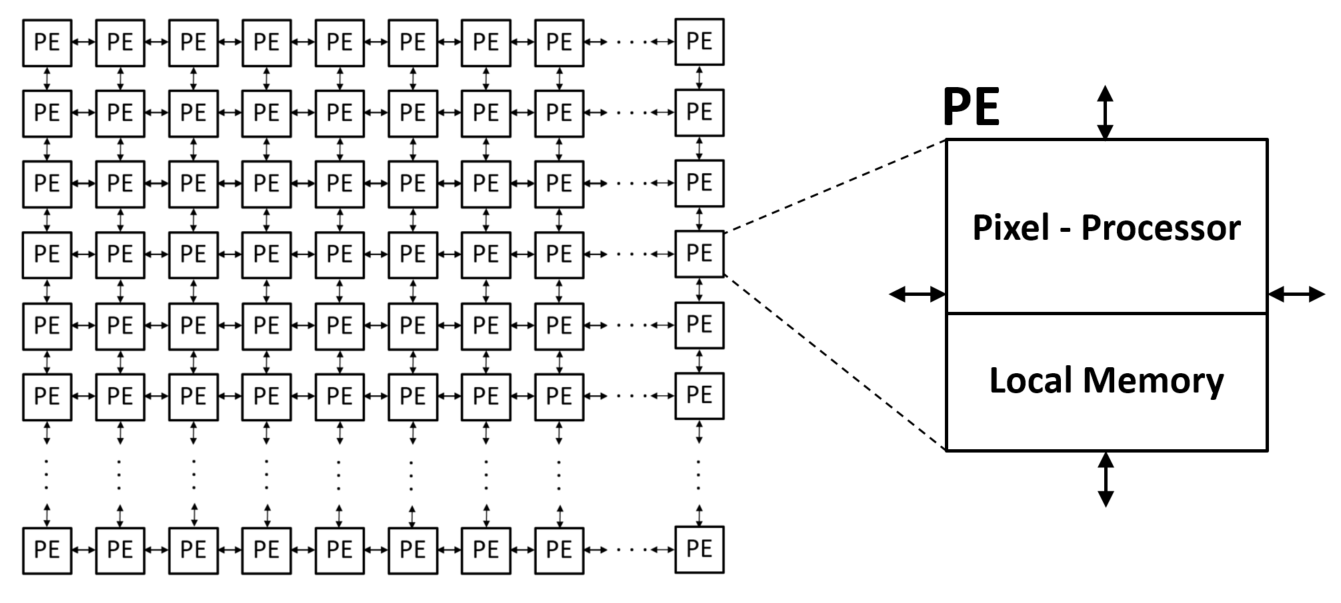}
\caption{\label{fig:PPA}A Pixel Processor Array contains a SIMD array of processing elements (PEs), on a 2D grid, where each PE contains processing and local memory and is allocated to one pixel in the image.}
\end{figure}

Recent trends in edge computing bring to the fore concerns about power efficiency of processing hardware. Some of the most challenging applications are in computer vision, where large amounts of raw sensory data (image pixels) need to be processed. It is well known that data movements are currently the most critical operations, responsible for energy consumption as well as the overall speed of the system. Minimising external memory access has become a necessity, and one of the solutions is a distributed architecture, with memory and processing resources collocated on a single device. As many low-level image processing tasks are inherently parallel, with identical operations executed for all pixels in the image, they are well suited to massively parallel SIMD (Single Instruction Multiple Data) architectures. An extreme level of parallelism can be achieved by allocating a processor per pixel, in a fine-grained SIMD architecture (Figure \ref{fig:PPA}). A very large number of processing elements, each containing local memory and arithmetic logic units, can efficiently execute pixel-parallel algorithms. Such cellular processor architectures have been considered in the past \cite{duff1978review, gealow1996system, ishikawa1999cmos, dudek2001general}. Thanks to recent advances in silicon fabrication technologies it is now possible to integrate thousands of elementary processors on a silicon chip, in a pixel-parallel image processor. Furthermore, it is now possible to integrate image sensing elements within the compute-memory fabric of the processor array on a single “vision chip” \cite{poikonen2009mipa4k,lopich2013aspa2,rodriguez2018cmos}. The co-location of photosensors and processors minimises the sensor-processor communications, providing additional benefits in terms of speed and power consumption of the system. We term such a device a Pixel Processor Array (PPA), where sensing, processing, and local memory are collocated on a processor-per-pixel basis. PPA vision sensors have been demonstrated, with resolutions up to $256 \times 256$ pixels \cite{carey2013100}. The recent technological trends of 3D silicon wafer stacking provide a vehicle for vertically integrating sensor and processor layers, promising future high-resolution vision sensor devices, where computing power can be placed behind each pixel of the image sensor \cite{yamazaki20174, millet20195500, finateu20205}.

The key advantage of PPA systems is that all low-level image processing occurs on the vision sensor integrated circuit, with no images transmitted off-chip in normal operation. Instead, only results of computations, for instance extracted features \cite{chen2017feature}, classification results \cite{Bose_2019_ICCV}, or visual odometry information \cite{bose2017visual}, are read-out directly from the device. To ensure that only low-dimensional data is read from the device, the PPA must perform all low-level image processing operations within the pixel-parallel array before employing sparse or summative read-out mechanisms. Many commonly used image processing operations, for instance local brightness adaptation, corner extraction, image convolution, etc. involve pixel-wise, localised computations. It is clear such operations are suited to PPA devices, however for many other tasks it is not obvious how mapping onto pixel-parallel architectures is achievable. In this paper we address the task of image transformations, illustrating how image rotation and scaling can be efficiently implemented on a pixel-parallel device.

The algorithms we propose are generally applicable to PPA devices, but in our implementation and experiments we use the SCAMP-5 vision sensor device \cite{carey2013100}. The architecture of the chip is briefly presented in the next section. Section \ref{sec:shearing} will introduce image shear operation which is then used to implement rotations described in Section \ref{sec:image rotation}. Section \ref{sec:scaling} will present the image scaling algorithm.

\section{SCAMP-5 Architecture \label{Sec:SCAMP-5}}
\begin{figure}
\centering
\includegraphics[width=1\linewidth]{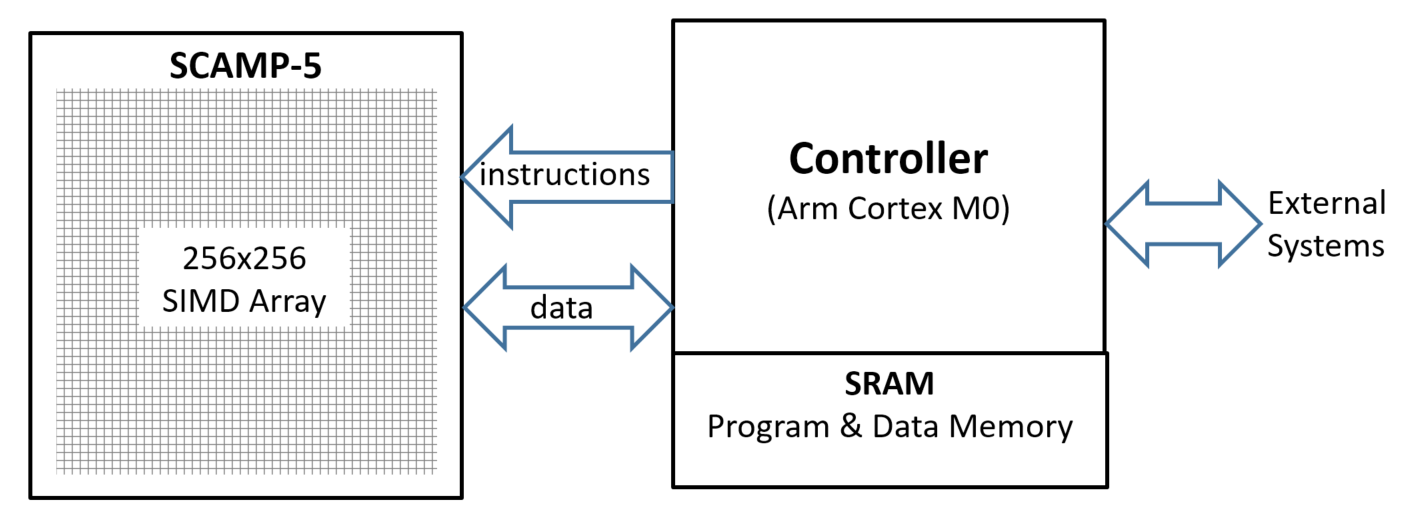}
\caption{\label{fig:SCAMP_array}Overview of the SCAMP vision system. The control program is executed on the ARM M0 core, which instructs the SCAMP-5 massively-parallel SIMD processor array to carry out operations on image arrays. SCAMP-5 has 256x256 Processing Elements.}
\end{figure}

The overall architecture of the hardware system used in this work is illustrated in Figure \ref{fig:SCAMP_array}. The SCAMP-5 chip comprises a $256 \times 256$ array of Processing Elements (PEs), which receive instructions from a single Controller (Arm Cortex-M0). The controller has its own program and data memory, and is responsible for the overall program flow, and any sequential computing required in the algorithm. It also issues microinstructions to the SCAMP-5 array. All PEs in the array execute the same microinstruction, issued by the Controller, i.e. the array operates as a SIMD processor.

Although it is possible to transfer data from the Controller to the SCAMP-5 array, the primary input to the array is optical, via photosensors in each PE. The typical operation is to acquire an image, and then process it in the SCAMP-5 array, according to the sequence of microinstructions sent by the Controller. The results of computations are read-out from the SCAMP-5 array by the Controller. While reading out entire data arrays is possible (and useful for debugging purposes), the fundamental read-out mechanisms is a sparse, “address-event” type of read-out. As a result of processing, the images are reduced to binary maps, preferably containing only a few non-zero pixels, and the row-column addresses of these pixels are sequentially extracted by the SCAMP-5 readout hardware, so that the array information is reduced to a few 16-bit addresses only. 

\begin{figure}
\centering
\includegraphics[width=1\linewidth]{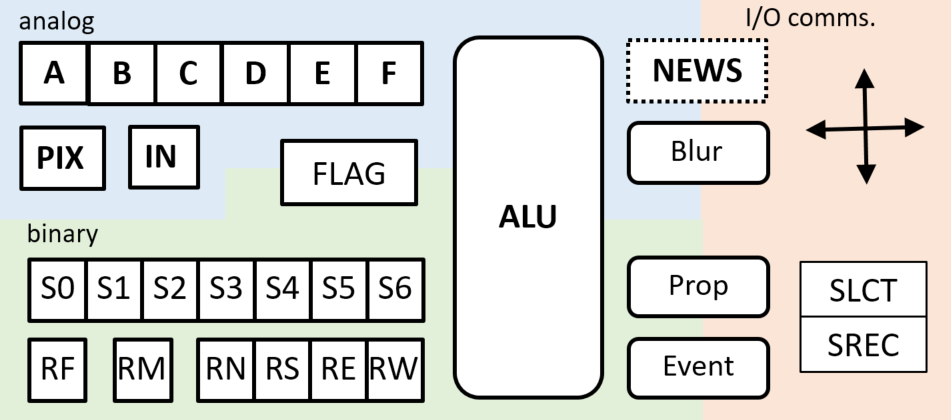}
\caption{\label{fig:SCAMP_PE}The architecture of the SCAMP-5 Processing Element. A-F are analog registers, PIX is image sensor input, IN is a global input. S0-S6 are general-purpose binary registers. Rx are special-purpose registers. ALU executes transfers and arithmetic and logic operations, 'Blur' and 'Prop' are  additional asynchronous hardware accelerators. FLAG is local activity register. NEWS provides 4-neighbour communications. SLCT and SREC provide array addressing and 'Event' unit enables sparse read-out. }
\end{figure}

The detail of the PE architecture is shown in Figure \ref{fig:SCAMP_PE}. Each PE contains six general-purpose “analog” registers that can store a gray-level pixel value or results of arithmetic operations, and thirteen binary registers. Several binary registers also have special-purpose designations. The ALU provides basic arithmetic and logic operations on the registers, for instance addition or subtraction of two analog registers, or logic AND operation on binary registers.

The FLAG register is a binary activity flag, used to implement conditional instruction execution. In each PE this can be set or reset individually, providing a degree of local autonomy. Only the PEs with FLAG set will execute SIMD instructions issued by the controller, otherwise these instructions are ignored.

The NEWS register is used to provide a mechanism for transferring data between a PE and its four nearest neighbours in the array. For instance, it is possible to move content of register A, to the same register A located in the PE’s neighbour to the South. From the point of view of the PE array, this results in a shift of data one pixel to the South. The transfer in binary registers is achieved using a multi-directional propagation operation, with the direction of transfer controlled by additional registers, for instance by setting RN=1 and RS=1, the operation S0=DNEWS(S0) will result in the value of S0 propagating simultaneously in both vertical directions.

The details of the SCAMP-5 implementation can be found in \cite{carey2013100}. The datapath is implemented using mixed-signal circuits, in particular storage and arithmetic operations on registers A-F are using analog current-mode
signal representation. This has some implications with respect to the precision and accuracy of arithmetic operations, and often requires special care be taken to ensure the inherent processing errors do not adversely affect the computation results. These considerations are beyond the scope of this paper. In many situations, processors can be programmed with the assumption that computations on registers A-F are roughly equivalent to 8-bit accuracy.

The analog current-mode computations allow operations such as global summation (all elements of the array are effectively added in one clock cycle) but this has limited precision.

When implementing vision algorithms on this architecture, a challenge is how to map the required image processing operations onto the constrained processor hardware. Currently, the SCAMP-5 array is programmed using in-line assembler code, while the overall Controller code can be compiled using standard C/C++. The work on more sophisticated compilation tools for this system is on-going \cite{martel2016vision, debrunner2019auke}.

Another challenge, is how to map the required computations onto the pixel-per-processor topology. This is illustrated by the algorithms introduced in the following sections, which demonstrate how pixel-parallel operations can be used to perform non-trivial image transformations such as rotations and scaling.

\begin{figure}
\centering
\includegraphics[width=0.99\linewidth]{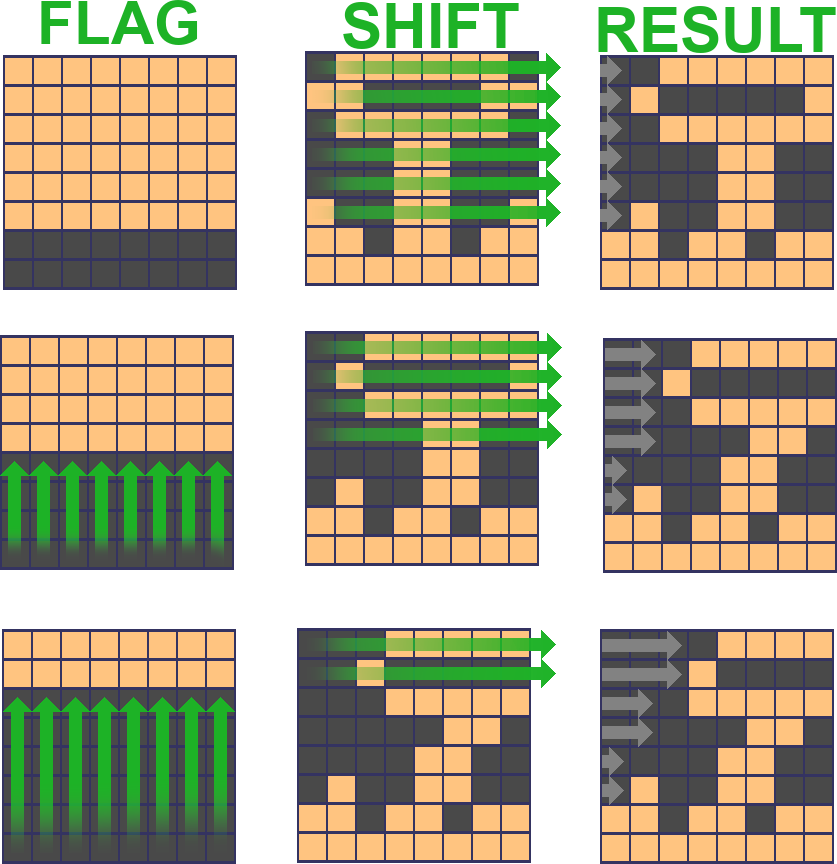}
\caption{Illustration of performing three steps of a horizontal shear. The FLAG register (Left Column) determines along which PE rows data is shifted at each stage. The FLAG register content itself is also shifted upwards in-between each step. As data is repeatedly shifted along the flagged rows the image becomes sheared (Top-Right to Bottom-Right). \label{fig:shear diagram}}
\end{figure}

\section{Shear Transformations \label{sec:shearing}}

A 2D shear transformation shifts all points parallel to some line through the origin.
For each point the direction and magnitude of this shift is proportional to its signed distance from said line. 
In this work we only consider shears parallel to the X and Y axi, which when combined correctly can be used to form various other transformations as demonstrated later in Section \ref{sec:image rotation}.
The matrices for shear transformations parallel to the X and Y axi respectively are given in Equations \ref{eq: shear x} and \ref{eq: shear y} respectively.

\begin{equation}
    \begin{pmatrix}
    x^{'}\\ 
    y^{'}
    \end{pmatrix}
    =
    \begin{pmatrix}
    1 & \alpha \\
    0 & 1
    \end{pmatrix}
    \begin{pmatrix}
    x\\ 
    y
    \end{pmatrix}
    =
    \begin{pmatrix}
    x +\alpha y\\ 
    y
    \end{pmatrix}
    \label{eq: shear x}
\end{equation}

\begin{equation}
    \begin{pmatrix}
    x^{'}\\ 
    y^{'}
    \end{pmatrix}
    =
    \begin{pmatrix}
    1 & 0 \\
    \alpha & 1
    \end{pmatrix}
    \begin{pmatrix}
    x\\ 
    y
    \end{pmatrix}
     =
    \begin{pmatrix}
    x \\ 
    y+\alpha x
    \end{pmatrix}
    \label{eq: shear y}
\end{equation}

The shear matrix of Equation \ref{eq: shear x} performs a one-to-one mapping, taking each point $(x,y)$ to its shifted position $(x+\alpha y,y)$. In this case each point's X coordinate is altered proportional to its Y coordinate, constituting a horizontal shear parallel to the X axis. Similarly, a vertical shear parallel to the Y axis, as defined in Equation \ref{eq: shear y}, operates in a similar manner.

\begin{figure*}
\centering
\includegraphics[width=0.24\linewidth]{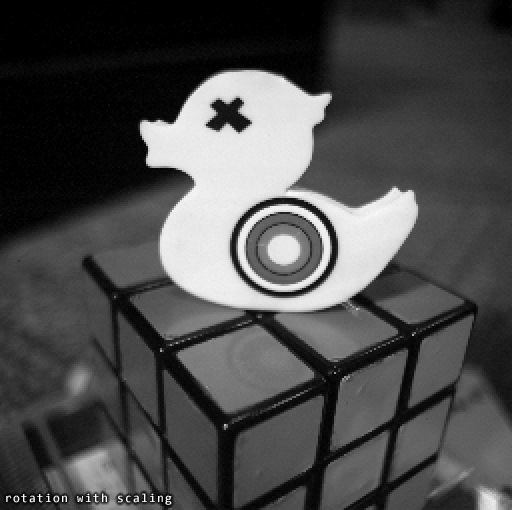}
\includegraphics[width=0.24\linewidth]{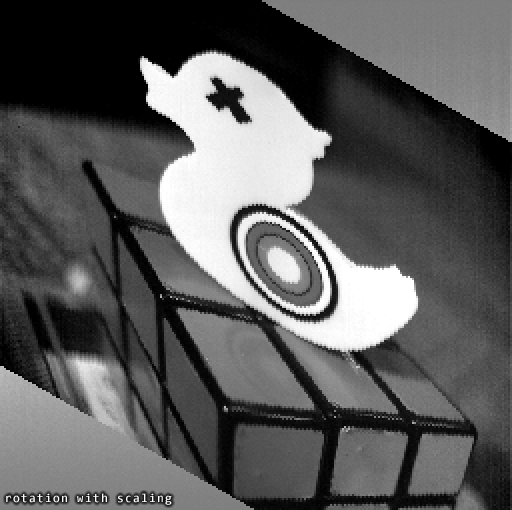}
\includegraphics[width=0.24\linewidth]{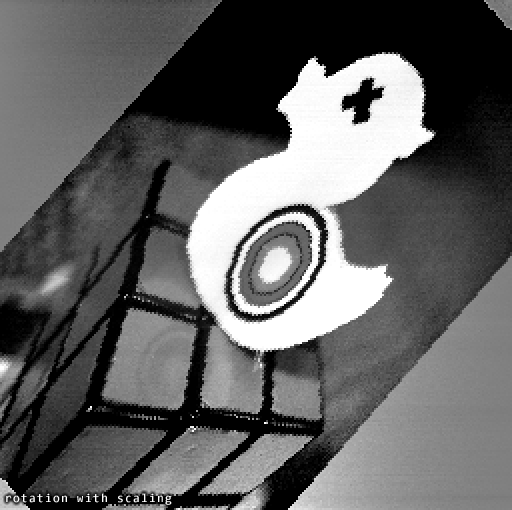}
\includegraphics[width=0.24\linewidth]{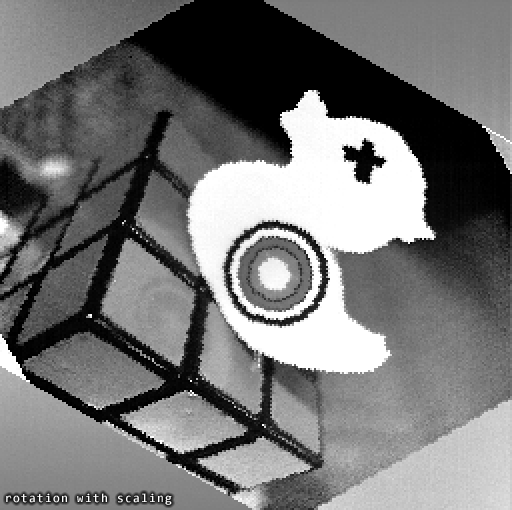}
\caption{\label{fig:three_shear_rot}Example of performing three consecutive shear operations upon an image stored upon SCAMP-5, resulting in an image rotation.}
\end{figure*}

\subsection{Shearing Upon SCAMP-5 PE Array}

A standard image shearing operation is performed by effectively shifting each pixel to a new location upon the 2D plane, and then determining a new image from these shifted pixels.
This new image is composed of a grid of pixels whose values are formed by interpolating between the values of the shifted pixels.
However performing interpolation is challenging upon current SCAMP-5 hardware, and so we instead consider only nearest neighbour image transformations.
In a nearest neighbor shear transformation, each pixel in the new image is a direct copy of the nearest shifted pixel.
For images stored upon the registers of the SCAMP-5's processor array, this translates to having to move the Pixel data stored within each PE to a new location on the array, that being the PE closest to the Pixel's shifted position.

As described in Section \ref{Sec:SCAMP-5}, each PE is only capable of directly transferring data to its immediate neighbours in the processor array.
Data can still be transferred between any two PEs indirectly by performing a sequence of data transfers, shuffling data from one PE to the next across the array, until it has been copied into the desired PE.
This data transfer is performed in parallel across all PEs in the array, however the FLAG register in each PE for conditional execution of SIMD instructions can be used to restrict data transfer operation to only select PEs.

It should be clear given these capabilities that image shearing transformations upon SCAMP-5 should be possible, the question remains how to perform a shear efficiently by exploiting the parallel compute of the SCAMP-5.

\subsection{X and Y Shearing Method}

Consider the shear parallel to the X axis as given by Equation \ref{eq: shear x}, horizontally shifting each row of pixels by an amount proportional to the row's Y location.
Conducting the equivalent nearest neighbour shear operation upon SCAMP-5 would involve repeated data transfers between PEs, shifting stored pixel data horizontally across the processor array.

This horizontal shift of data could be performed one PE row at a time, however this would be slow and highly inefficient.
Instead our proposed approach horizontally shifts the pixel data stored across of multiple rows of PEs simultaneously in parallel.

Note again that the SCAMP-5 PPA consists of a $256 \times 256$ array of PEs.
Taking the origin of a stored image to be at the center of the PE array, when performing a horizontal shear such as in Equation \ref{eq: shear x}, the $i^{th}$ row of PEs will require a horizontal shift $r_{i}$ as given by Equation \ref{eq: row shift}.

\begin{equation}
r_{i} = Ceil(\alpha(128-i))  
\label{eq: row shift}
\end{equation}{}
With data in the top and bottom halves of the array being shifted in opposite directions, and with shifts up to a magnitude of $N = Ceil(\mid \alpha 128 \mid)$.
Let $S_{n}$ denote the set of indices for all rows that require a shift of $n$ as shown in Equations \ref{eq: row shift set pos}.
\begin{equation}
S_{n} = \{ i \mid r_{i}= n \in \mathbb{Z} \}
\label{eq: row shift set pos}
\end{equation}{}
Let us assume that $\alpha > 0$, and hence all PE rows in the top half of the array belong to one of the sets $S_{1},S_{2},S_{3}...S_{N}$.
To efficiently conduct the shear transformation, the pixel data of these PE rows should be shifted using as few parallel data transfer operations as possible.
This can be achieved by performing a single horizontal data transfer operation upon the PE rows of $S_{1} \cup S_{2} \cup S_{3} \cup...S_{N}$, then repeating on the rows of $S_{2} \cup S_{3} \cup S_{4} \cup...S_{N}$, then $S_{3} \cup S_{4} \cup S_{5} \cup...S_{N}$ and so on. 
Doing so shifts pixel data across multiple rows simultaneously, with each row stopping once its data is correctly shifted according to the shear transformation being performed.

In practice, performing this coordinated data shifting across PE rows requires correct manipulation of the PE FLAG registers controlling conditional execution of SIMD instructions.
In the top half of the array the distance to shift each PE row increases going upwards, meaning the FLAG registers within PE rows must be toggled off successively from bottom to top.
When viewed as an image, the content of the FLAG registers will then appear as a sweeping curtain, moving upward with each successive horizontally shift as is illustrated in Figure \ref{fig:shear diagram}.
Such a sweeping curtain moving across the FLAG registers can be efficiently created using data transfer operations to shift each PEs FLAG register content upwards.

Once this coordinated shifting of data has been performed for the top half of the PE array, a similar routine can be performed for the bottom half completing the shear operation by transferring all stored pixel into the correct PE locations.
The method for this approach is laid out in [\ref{alg:horizontal shear}], where Shift(X,DIR) denoted performing a parallel data transfer across all active flagged PEs, copying the content of register X into the same register of a neighbouring PE element.
Vertical shearing can be performed in the same manner but now splitting the PE array into left and right halves and vertically shifting columns of PEs in-place of rows.

\begin{algorithm}
\begin{algorithmic}[0] 
\State{$A$ \kern 3mm \textit{// register holding pixel data in each PE}}
\State{$N = Ceil(\mid\alpha128\mid$) \kern 3mm \textit{// Greatest Required Shift}}
\\
\State{\textit{//Flag all PEs in rows in top half to be shifted}}
\State{Clear\_FLAG (all PEs)}
\State{Set\_FLAG (PEs in rows from $0$ to $Max(S_{1})$)}
\\
\State{\textit{//Shift pixel data in top half of array}}
\For{$n = 1$ to $N$}
    \State{\textit{//Horizontally shift data in Flagged PEs}}
    \If{$\alpha > 0$}
          \State{Shift(A,WEST)}
    \Else{}
          \State{Shift(A,EAST)}
    \EndIf{}
    \State{\textit{//Vertically shift FLAG registers}}
    \State{$V = Max(S_{n+1})- Max(S_{n})$}
    \For{$i = 0$ to $V$}
         \State{Shift(FLAG,SOUTH)}
    \EndFor
\EndFor

\\
\State{\textit{//Flag all PEs in rows to be shifted}}
\State{Clear\_FLAG (all PEs)}
\State{Set\_FLAG (PEs in rows from $Min(S_{-1})$ to $255$)}
\\
\State{\textit{//Shift pixel data in top half of array}}
\For{$n = 1$ to $N$}
    \State{\textit{//Horizontally shift data in Flagged PEs}}
    \If{$\alpha > 0$}
          \State{Shift(A,EAST)}
    \Else{}
          \State{Shift(A,WEST)}
    \EndIf{}
    \State{\textit{//Vertically shift FLAG registers}}
    \State{$V = Min(S_{-n}) - Min(S_{-(n+1)})$}
    \For{$i = 0$ to $V$}
        \State{Shift(FLAG,NORTH)}
    \EndFor
\EndFor
\end{algorithmic}

\protect\caption{Horizontal Shearing 
\label{alg:horizontal shear}}
\end{algorithm}

\section{Rotation By Three Shears \label{sec:image rotation}}
To rotate images stored upon the SCAMP-5 we make use of the fact that any arbitrary rotation matrix of $\theta$ radians as in Equation \ref{eqn:rotmat}, can be decomposed into a combination of three shear matrices such as shown in Equation \ref{eqn:rotmatshears}. 
\begin{equation}
\begin{bmatrix}
cos(\theta) & sin(\theta) \\
-sin(\theta) & cos(\theta)
\end{bmatrix}
\label{eqn:rotmat}
\end{equation}

\begin{equation}
\begin{bmatrix}
    1 & -tan(\frac{\theta}{2}) \\
	0 & 1
\end{bmatrix}
\begin{bmatrix}
    1 & 0 \\
	sin(\theta) & 1
\end{bmatrix}
\begin{bmatrix}
    1 & -tan(\frac{\theta}{2}) \\
	0 & 1
\end{bmatrix}
\label{eqn:rotmatshears}
\end{equation}
In this case two horizontal shear operations and one vertical in the same form as described previously in Section \ref{sec:shearing}.
Performing such a set of three shearing operations in sequence will then result in the pixel data being shifted across the PEs of the array such that the appropriate rotated image is produced.
An example of such a rotation being performed one shear at a time upon SCAMP-5 is illustrated in Figure \ref{fig:three_shear_rot}.
The time required to perform such a rotation linearly increases with the rotation angle, with a large rotation of $45$ degrees taking $1031 \mu s$ to perform.

\section{Image Scaling \label{sec:scaling}}

\begin{figure*}
\centering
\includegraphics[width=0.24\linewidth]{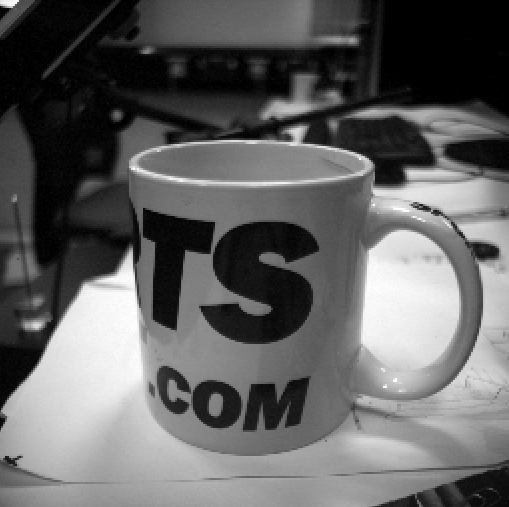}
\includegraphics[width=0.24\linewidth]{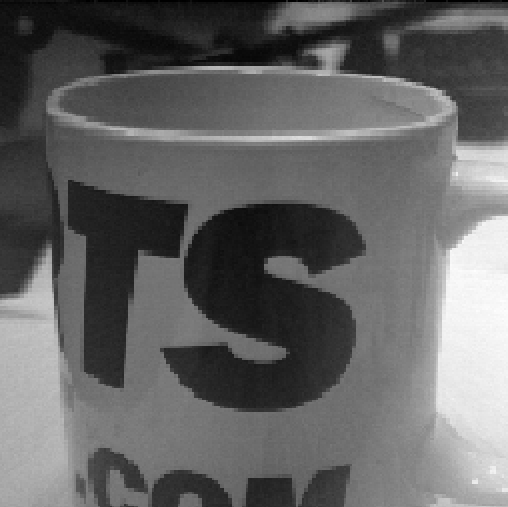}
\includegraphics[width=0.24\linewidth]{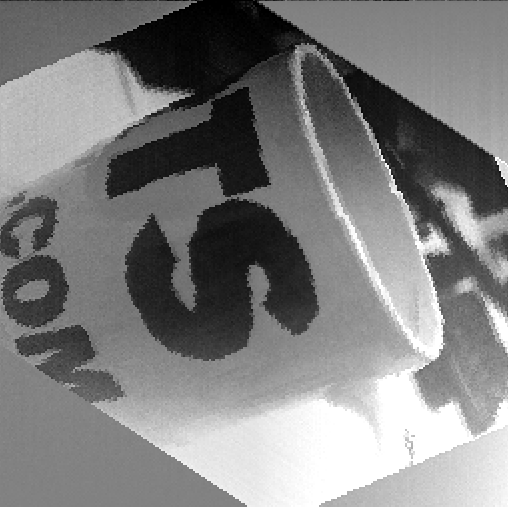}
\includegraphics[width=0.24\linewidth]{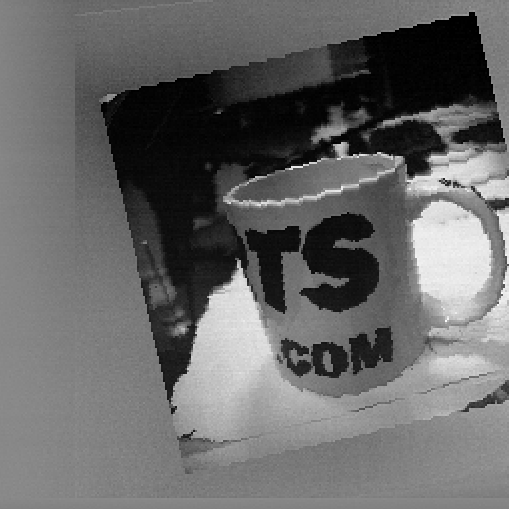}
\caption{\label{fig:image_scaling}Examples of in-plane image transformations performed on SCAMP-5. Top Left:original, Top Right:Up-scaled, Bottom Left:Up-scaled and Rotated, Bottom right:Down-scaled and Rotated}
\end{figure*}

A typical scaling transformation moves each point $(x,y)$ to a new location $(\alpha x, \beta y)$, bringing said point either closer or further from each of the X and Y axi depending upon scaling factors $\alpha$ and $\beta$.
Here however we consider how to perform the separate cases of horizontal and vertical scaling operations as shown in Equations \ref{eq:scaling x}, \ref{eq:scaling y} upon SCAMP-5.
Such horizontal and vertical scaling operations can then be performed in sequence to perform any general image scaling as illustrated in the examples of Figure \ref{fig:image_scaling}.

\begin{equation}
    \begin{pmatrix}
    x^{'}\\ 
    y^{'}
    \end{pmatrix}
    =
    \begin{pmatrix}
    \alpha & 0 \\
    0 & 1
    \end{pmatrix}
    \begin{pmatrix}
    x\\ 
    y
    \end{pmatrix}
     =
    \begin{pmatrix}
    \alpha x \\ 
     y 
    \end{pmatrix}
    \label{eq:scaling x}
\end{equation}

\begin{equation}
    \begin{pmatrix}
    x^{'}\\ 
    y^{'}
    \end{pmatrix}
    =
    \begin{pmatrix}
    1 & 0 \\
    0 & \beta
    \end{pmatrix}
    \begin{pmatrix}
    x\\ 
    y
    \end{pmatrix}
     =
    \begin{pmatrix}
    x \\ 
    \beta y 
    \end{pmatrix}
    \label{eq:scaling y}
\end{equation}

Similar to the image shearing of Section \ref{sec:shearing}, we implement nearest neighbour image scaling due to the difficulties in implementing pixel interpolation upon SCAMP-5. 
Such nearest neighbour scaling involves eliminating or duplicating rows and/or columns of pixel data across the image.
Specifically horizontal down-scaling ($\alpha < 1$) involves eliminating columns of pixel data, while horizontal up-scaling ($\alpha > 1$) involves inserting duplicated columns of pixel data.
A similar process is performed for vertical scaling, but upon rows of stored pixel data instead.
In both cases we regard the origin of the image to be located at the center of the PE array.

\begin{figure}
\centering
\includegraphics[width=0.99\linewidth]{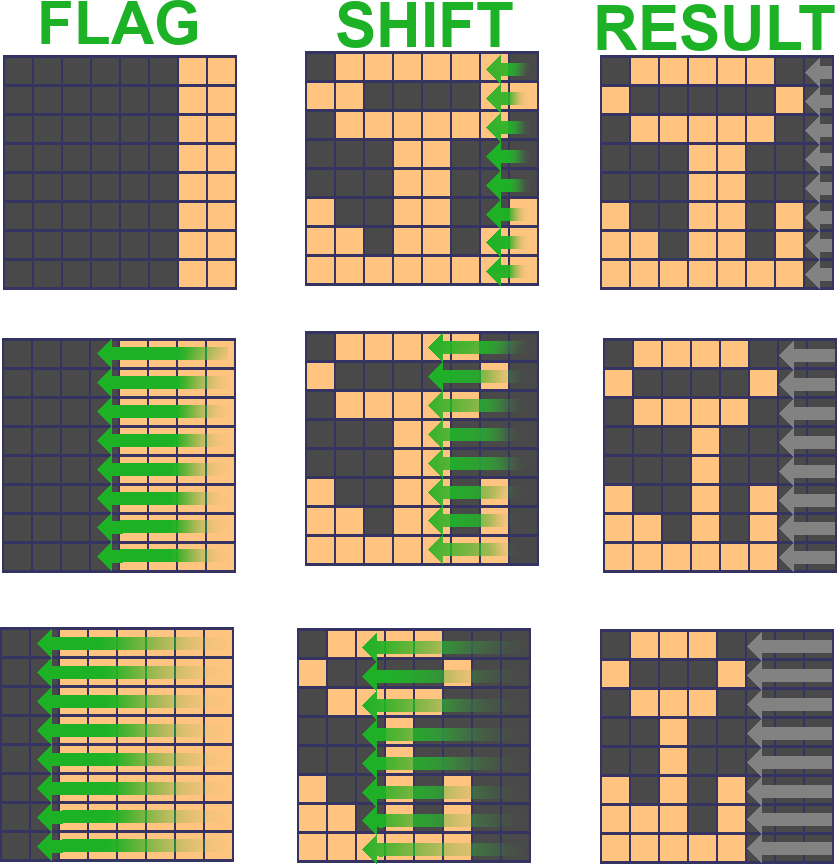}
\caption{Illustration of performing horizontal down-scaling over several steps (Top-Right to Bottom-Right). The FLAG register (Left Column) is used to select which columns of PEs should copy data from their rightmost neighbours. This eliminates one column of data in the process by overwriting its content, shrinking the remaining image. The FLAG register content itself is shifted in-between each step, selecting the next column to be eliminated. \label{fig:scale diagram}}
\end{figure}

\subsection{Scaling Upon SCAMP-5}

Duplication or elimination of a row/column of pixel data can be performed on SCAMP-5 using the parallel data transfer operation employed previously for image shearing. 
By correctly setting up the FLAG registers, limiting the data transfer into only select PEs, this operation can be used not only to shift pixel data but also overwrite it. 

Let us examine horizontal down-scaling of an analog image upon SCAMP-5 i.e. $0 < \alpha < 1$ (for $\alpha$ in Equation \ref{eq:scaling x}), by eliminating columns of pixel data.
As the image origin is taken to be the center of the array, this scaling will require horizontally shifting data in the left and right sides of the array in opposite directions, bringing data from both sides towards the array's center.
Further, to produce a correctly scaled image the columns eliminated from this data shifting must be evenly spaced across the array.
As the PE array is 256 elements in width, the number of columns to eliminate from both the left and right hand sides is given by $E$ in Equation \ref{eq: cols to remove}
\begin{equation}
E = 128-Ceil(\alpha 128)  
\label{eq: cols to remove}
\end{equation}{}
with even spacing between these eliminated columns then given by $K$ in Equation \ref{eq: col spacing}.
\begin{equation}
K = Ceil(128/E)  
\label{eq: col spacing}
\end{equation}{}
Let us examine only the right hand side of the array for now.
To eliminate the first column of pixel data from this side, the PE FLAG registers are set within that column along with all other columns to the right.
A parallel data transfer operation is then performed, instructing each flagged PE to copy over the data from the PE to its right.
This causes the pixel data across all flagged columns to be shifted to the left, except in the column to be eliminated, whose pixel data is then overwritten eliminating it from the image.

This process can then be repeated for all remaining columns that are to be eliminated from this side of the array.
However for each of these subsequent columns, the FLAG registers required to perform this elimination do not need to be setup from scratch.
Instead the FLAG register content used for the previous column elimination can itself be shifted horizontally to flag the necessary columns of PEs.
Thus, as columns are eliminated the FLAG register content takes the form of a sliding curtain, similar to that used in Section \ref{sec:shearing} for image shearing, providing an efficient means to repeatedly eliminate columns of pixel data from the array.
This approach is illustrated in Figure \ref{fig:scale diagram}.
The same approach can then be performed to eliminate columns from the left hand side of the image, completing the horizontal down-scaling as listed in Algorithm \ref{alg:horizontal down scaling}

Vertical down-scaling of an image can be performed in much of the same way, largely just switching the routine from using columns to rows.
Similarly up-scaling of an image is performed by in a highly similar fashion, except now duplicating rows or columns of pixel data at evenly spaced intervals, rather than eliminating them.
The time taken to perform such scaling operations increases with their magnitude, with up-scaling an image by a factor of 2 (scaling both in x and y) taking $445 \mu s$, and down-scaling to half size taking the same time.

\begin{algorithm}
\begin{algorithmic}[0] 
\State{$A$ \kern 3mm \textit{// register holding pixel data in each PE}}
\State{$K$ \kern 3mm \textit{// Column skip value determining down-scaling}}
\State{$N = Ceil(128/K)$ \kern 3mm \textit{//Required Shifts}}
\\
\State{\textit{//Setup FLAG for scaling right half of array}}
\State{Clear\_FLAG (all PEs)}
\State{Set\_FLAG (PEs in columns from $0$ to $K$)}
\\
\State{\textit{//Scale pixel data in right half of array}}
\For{$n = 1$ to $N$}
    \State{Shift(A,WEST) \kern 3mm \textit{// Shift data in Flagged PEs}}
    \For{$i = 0$ to $K$}
         \State{Shift(FLAG,WEST) \kern 3mm \textit{//Shift FLAG register content}}
    \EndFor
\EndFor
\\
\State{\textit{//Setup FLAG for scaling left half of array}}
\State{Clear\_FLAG (all PEs)}
\State{Set\_FLAG (PEs in columns from ($256-K$) to $K$)}
\\
\State{\textit{//Scale pixel data in left half of array}}
\For{$n = 1$ to $N$}
    \State{Shift(A,EAST) \kern 3mm \textit{// Shift data in Flagged PEs}}
    \For{$i = 0$ to $K$}
         \State{Shift(FLAG,EAST) \kern 3mm \textit{//Shift FLAG register content}}
    \EndFor
\EndFor
\end{algorithmic}

\protect\caption{Horizontal Down-Scaling 
\label{alg:horizontal down scaling}}
\end{algorithm}

\section{Links}

\begin{itemize}

\item\href{https://www.youtube.com/watch?v=OESElS0B6Vs}{Video Demonstration Of Transformations}.

\item\href{https://github.com/lauriebose/Scamp5-Image-Transformations}{Image Transformations Source Code - Scamp5}.

\item\href{https://github.com/lauriebose/Scamp7-Image-Transformations}{Image Transformations Source Code - Scamp7}.

\end{itemize}

\section{Conclusions}
This paper presented a set of novel algorithms for conducting image transformations upon PPA devices.
In each case we presented a new approach which exploits the parallel processing of the SCAMP-5, but which should also be applicable to PPA architectures in general.
Our implementations are fast enough to be used as essential standard functions in many real-time works.
It is our hope that this paper helps demonstrate how a wide range of tasks are possible PPA devices, and that others may use the work presented here to accelerate building their own applications.

{\small
\bibliographystyle{unsrt}
\bibliography{article}

\begin{thebibliography}{10}

\bibitem{duff1978review}
Michael~JB Duff et~al.
\newblock Review of the {CLIP} image processing system.
\newblock In {\em Proc. National Computer Conference}, pages 1055--1060. AFIPS
  Press Arlington, Va, 1978.

\bibitem{gealow1996system}
Jeffrey~C Gealow, Frederick~P Herrmann, Lawrence~T Hsu, and Charles~G Sodini.
\newblock System design for pixel-parallel image processing.
\newblock {\em IEEE Transactions on Very Large Scale Integration (VLSI)
  Systems}, 4(1):32--41, 1996.

\bibitem{ishikawa1999cmos}
Masatoshi Ishikawa, Kazuya Ogawa, Takashi Komuro, and Idaku Ishii.
\newblock A cmos vision chip with simd processing element array for 1ms image
  processing, 1999 dig. tech. papers of 1999 ieee int.
\newblock In {\em Solid-State Circuits Conf.(ISSCC99)(San Francisco, 1999.2.
  16)/Abst}, pages 206--207.

\bibitem{dudek2001general}
Piotr Dudek and Peter~J Hicks.
\newblock A general-purpose cmos vision chip with a processor-per-pixel simd
  array.
\newblock In {\em Proceedings of the 27th European Solid-State Circuits
  Conference}, pages 213--216. IEEE, 2001.

\bibitem{poikonen2009mipa4k}
Jonne Poikonen, Mika Laiho, and Ari Paasio.
\newblock {MIPA4k}: A 64$\times$ 64 cell mixed-mode image processor array.
\newblock In {\em 2009 IEEE International Symposium on Circuits and Systems},
  pages 1927--1930. IEEE, 2009.

\bibitem{lopich2013aspa2}
Alexey Lopich and Piotr Dudek.
\newblock A general-purpose vision processor with 160x80 pixel-parallel {SIMD}
  processor array.
\newblock In {\em Proceedings of the IEEE Custom Integrated Circuits
  Conference}, 2017.

\bibitem{rodriguez2018cmos}
A~Rodriguez-Vazquez, Jorge Fern{\'a}ndez-Berni, Juan~Antonio
  Le{\~n}ero-Bardallo, I~Vornicu, and Ricardo Carmona-Gal{\'a}n.
\newblock {CMOS} vision sensors: embedding computer vision at imaging
  front-ends.
\newblock {\em IEEE Circuits and Systems Magazine}, 18(2):90--107, 2018.

\bibitem{carey2013100}
Stephen~J Carey, Alexey Lopich, David~RW Barr, Bin Wang, and Piotr Dudek.
\newblock A 100,000 fps vision sensor with embedded {535GOPS/W} 256$\times$ 256
  {SIMD} processor array.
\newblock In {\em 2013 Symposium on VLSI Circuits}, pages C182--C183. IEEE,
  2013.

\bibitem{yamazaki20174}
Tomohiro Yamazaki, Hironobu Katayama, Shuji Uehara, Atsushi Nose, Masatsugu
  Kobayashi, Sayaka Shida, Masaki Odahara, Kenichi Takamiya, Yasuaki Hisamatsu,
  Shizunori Matsumoto, et~al.
\newblock A 1ms high-speed vision chip with 3d-stacked 140 {GOPS}
  column-parallel {PE}s for spatio-temporal image processing.
\newblock In {\em 2017 IEEE International Solid-State Circuits Conference
  (ISSCC)}, pages 82--83. IEEE, 2017.

\bibitem{millet20195500}
Laurent Millet, Stephane Chevobbe, Caaliph Andriamisaina, Lamine Benaissa,
  Edouard Deschaseaux, Edith Beigne, Karim~Ben Chehida, Maria Lepecq, Mehdi
  Darouich, Fabrice Guellec, et~al.
\newblock A 5500-frames/s 85-{GOPS/W} 3-d stacked {BSI} vision chip based on
  parallel in-focal-plane acquisition and processing.
\newblock {\em IEEE Journal of Solid-State Circuits}, 54(4):1096--1105, 2019.

\bibitem{finateu20205}
Thomas Finateu, Atsumi Niwa, Daniel Matolin, Koya Tsuchimoto, Andrea
  Mascheroni, Etienne Reynaud, Pooria Mostafalu, Frederick Brady, Ludovic
  Chotard, Florian LeGoff, et~al.
\newblock A 1280$\times$ 720 back-illuminated stacked temporal contrast
  event-based vision sensor with 4.86 $\mu$m pixels, 1.066 {GEPS} readout,
  programmable event-rate controller and compressive data-formatting pipeline.
\newblock In {\em 2020 IEEE International Solid-State Circuits
  Conference-(ISSCC)}, pages 112--114. IEEE, 2020.

\bibitem{chen2017feature}
Jianing Chen, Stephen~J Carey, and Piotr Dudek.
\newblock Feature extraction using a portable vision system, 2017.

\bibitem{Bose_2019_ICCV}
Laurie Bose, Jianing Chen, Stephen~J. Carey, Piotr Dudek, and Walterio
  Mayol-Cuevas.
\newblock A camera that cnns: Towards embedded neural networks on pixel
  processor arrays.
\newblock In {\em The IEEE International Conference on Computer Vision (ICCV)},
  October 2019.

\bibitem{bose2017visual}
Laurie Bose, Jianing Chen, Stephen~J Carey, Piotr Dudek, and Walterio
  Mayol-Cuevas.
\newblock Visual odometry for pixel processor arrays.
\newblock In {\em Proceedings of the IEEE International Conference on Computer
  Vision}, pages 4604--4612, 2017.

\bibitem{martel2016vision}
Julien~NP Martel and Piotr Dudek.
\newblock Vision chips with in-pixel processors for high-performance low-power
  embedded vision systems.
\newblock In {\em ASR-MOV Workshop, CGO}, volume~6, page~14, 2016.

\bibitem{debrunner2019auke}
Thomas Debrunner, Sajad Saeedi, and Paul~HJ Kelly.
\newblock {AUKE}: Automatic kernel code generation for an analogue {SIMD}
  focal-plane sensor-processor array.
\newblock {\em ACM Transactions on Architecture and Code Optimization (TACO)},
  15(4):59, 2019.

\end{thebibliography}
}

\end{document}